\documentclass{article}
\usepackage{spconf,amsmath,graphicx}
\usepackage{xcolor,adjustbox,xspace}
\usepackage{multirow}
\usepackage{hyperref}
\hypersetup{
    colorlinks=true,
}
\usepackage[numbers]{natbib}

\usepackage{amssymb}

\usepackage{amsmath,amsfonts}
\usepackage{amsthm,amsopn}

\usepackage{bm} 

\newcommand{\vct}[1]{\boldsymbol{#1}} 

\newcommand{\field}[1]{\mathbb{#1}}
\newcommand{\R}{\field{R}} 
\newcommand{\T}{^{\textrm T}} 

\newcommand{\ProbOpr}[1]{\mathbb{#1}}

\newcommand{\expect}[2]{%
\ifthenelse{\equal{#2}{}}{\ProbOpr{E}_{#1}}
{\ifthenelse{\equal{#1}{}}{\ProbOpr{E}\left[#2\right]}{\ProbOpr{E}_{#1}\left[#2\right]}}} 
\newcommand{\var}[2]{%
\ifthenelse{\equal{#2}{}}{\ProbOpr{VAR}_{#1}}
{\ifthenelse{\equal{#1}{}}{\ProbOpr{VAR}\left[#2\right]}{\ProbOpr{VAR}_{#1}\left[#2\right]}}} 

\newcommand{\cov}[2]{%
\ifthenelse{\equal{#2}{}}{\mathrm{Cov}_{#1}}
{\ifthenelse{\equal{#1}{}}{\mathrm{Cov}\left[#2\right]}{\mathrm{Cov}_{#1}\left[#2\right]}}} 

\newcommand{\va}{\vct{a}}

\newcommand{\vf}{\vct{f}}

\newcommand{\vh}{\vct{h}}

\newcommand{\vv}{\vct{v}}
\newcommand{\vw}{\vct{w}}
\newcommand{\vx}{{\vct{x}}}

\newcommand{\vz}{{\vct{z}}}

\newcommand{\WA}{\textsf{WA}\xspace}
\newcommand{\UA}{\textsf{UA}\xspace}

\newcommand{\RNNT}{$\mathsf{RNN}$-$\mathsf{T}$\xspace} 
\newcommand{\TODO}[1]{{\color{blue} TODO: #1}}

\newcommand{\llcao}[1]{{\color{red} llcao: #1}}

\newcommand{\ie}{{i.e.}\ }

\newcommand{\eat}[1]{}

\title{SPEECH SENTIMENT ANALYSIS VIA PRE-TRAINED FEATURES \\ 
FROM END-TO-END ASR MODELS}
\name{Zhiyun Lu$^{1}$\sthanks{Work was done while the author interned at Google.}, Liangliang Cao$^{2}$, Yu Zhang$^{2}$, Chung-Cheng Chiu$^{2}$,  James Fan$^{2}$}
\address{
    $^1$University of Southern California, USA\hspace{3ex}
    $^2$Google, Inc., USA \\
    \fontsize{9}{9}\selectfont\ttfamily\upshape
    {zhiyunlu@usc.edu, \{llcao, ngyuzh, chungchengc, jjfan\}@google.com}
    \vspace{-0.2in}
    }

\begin{document}
%
\maketitle
%
\begin{abstract}
In this paper, we propose to use pre-trained features from end-to-end ASR models to solve speech sentiment analysis as a down-stream task. 
We show that end-to-end ASR features, which integrate both acoustic and text information from speech, achieve promising results. We use RNN with self-attention as the sentiment classifier, which also provides an easy visualization through attention weights to help interpret model predictions.
We use well benchmarked IEMOCAP dataset and a new large-scale speech sentiment dataset SWBD-sentiment for evaluation. Our approach improves the-state-of-the-art accuracy on IEMOCAP from 66.6\% to 71.7\%, and achieves an accuracy of 70.10\% on SWBD-sentiment with more than 49,500 utterances. 

\end{abstract}


\eat{the benefits from  from both acoustic models and language models. 
From the sequence of ASR features, we develop effective methods to recognize sentiment and get promising results.

which is created by adding sentiment labels to the Switchboard dataset, are used. 
}

\section{INTRODUCTION}
\label{sec:intro}

Speech sentiment analysis is an important problem for interactive intelligence systems with broad applications in many industries, e.g., customer service, health-care, and education. The task is to classify a speech utterance into one of a fixed set of categories, such as positive, negative or neutral. Despite its importance, speech sentiment analysis remains a challenging problem, due to rich variations in speech, like different speakers and acoustic conditions. In addition, existing sentiment datasets are relatively small-scale, which has limited the research development.

\eat{has been dramatically limited by the lack of large-scale
resources. }

The key challenge in speech sentiment analysis is how to learn a good representation that captures the emotional signals and remains invariant under different speakers, acoustic conditions, and other natural speech variations. 
Traditional approaches employed acoustic features, such as band-energies, filter banks, and MFCC features~\cite{li2019dilated, li2018attention, wu2019speech, xie2019speech}, or raw waveform~\cite{tzirakis2018end} to predict sentiment. 
However, models trained on low-level features can easily overfit to noise or sentiment irrelevant signals. One way to remove variations in speech is to transcribe the audio into text, and use text features to predict sentiment~\cite{lakomkin2019incorporating}. Nonetheless, sentiment signals in the speech, like laughter, can be lost in the transcription. 
Latest works~\cite{kim2019dnn, prabhavalkar2017comparison,cho2018deep,gu2018multimodal} try to combine acoustic features with text, but it is unclear what is the best way to fuse the two modalities. \eat{Additionally, since  n-gram models and text embeddings can bring millions of parameters, the combined model may easily overfit and do not generalize well in real applications. }
Other general feature learning techniques, like unsupervised learning~\cite{eskimez2018unsupervised} and multi-task learning~\cite{zhang2019attention} have also been explored.

In this work, we introduce a new direction to tackle the challenge. We propose to use end-to-end (e2e) automatic speech recognition (ASR)~\cite{prabhavalkar2017comparison,rnnt-asru17,chiu2018state,he2019streaming} as pre-training, and solve the speech sentiment as a down-stream task. 
This approach is partially motivated by the success of pre-training in solving tasks with limited labeled data in both computer vision 
and language. 
Moreover, the e2e model combines both acoustic and language models of traditional ASR, thus can seamlessly integrate the acoustic and text features into one representation. We hypothesize that the ASR pre-trained representation works well on sentiment analysis.
We compare different sentiment decoders on ASR features, and apply spectrogram augmentation~\cite{park2019specaugment} to reduce overfitting.

\eat{There is a growing interest in learning e2e ASR models recently, which have achieved state-of-the-art performances in a number of tasks. In this work, we want to further demonstrate the power of end-to-end models for down-stream tasks.} 

To further advance the study of the speech sentiment task, we annotate a subset of switchboard telephone conversations~\cite{godfrey1997switchboard} with sentiment labels, and create the SWBD-sentiment dataset\footnote{SWBD-sentiment will be released through \href{http://www.ldc.upenn.edu/}{Linguistic Data Consortium}.}. It contains 140 hours speech with 49,500 labeled utterances, which is 10 times larger than IEMOCAP~\cite{busso2008iemocap}, the current largest one. We evaluate the performance of pre-trained ASR features on both IEMOCAP and SWBD-sentiment. On IEMOCAP, we improve the state-of-the-art sentiment analysis accuracy from 66.6\% to 71.7\%. On SWBD-sentiment, we achieve 70.10\% accuracy on the test set, outperforming strong baselines.

\eat{

To illustrate these challenges, let's take a look at 
IEMOCAP~\cite{busso2008iemocap}, a well-known benchmark for speech sentiment analysis. It has been studied by many research works \cite{li2019dilated, cho2018deep, kim2019dnn,gu2018multimodal}, and  the state-of-the-art of accuracy at recognizing sentiments from IEMOCAP is 68.7\%. However, IEMOCAP only provides 12 hours of audios performed by professional actors, which may not fully represent the difficulty of sentiment analysis. 
}

\begin {figure}[!tbp]
\centering
    \includegraphics[width=0.5 \textwidth]{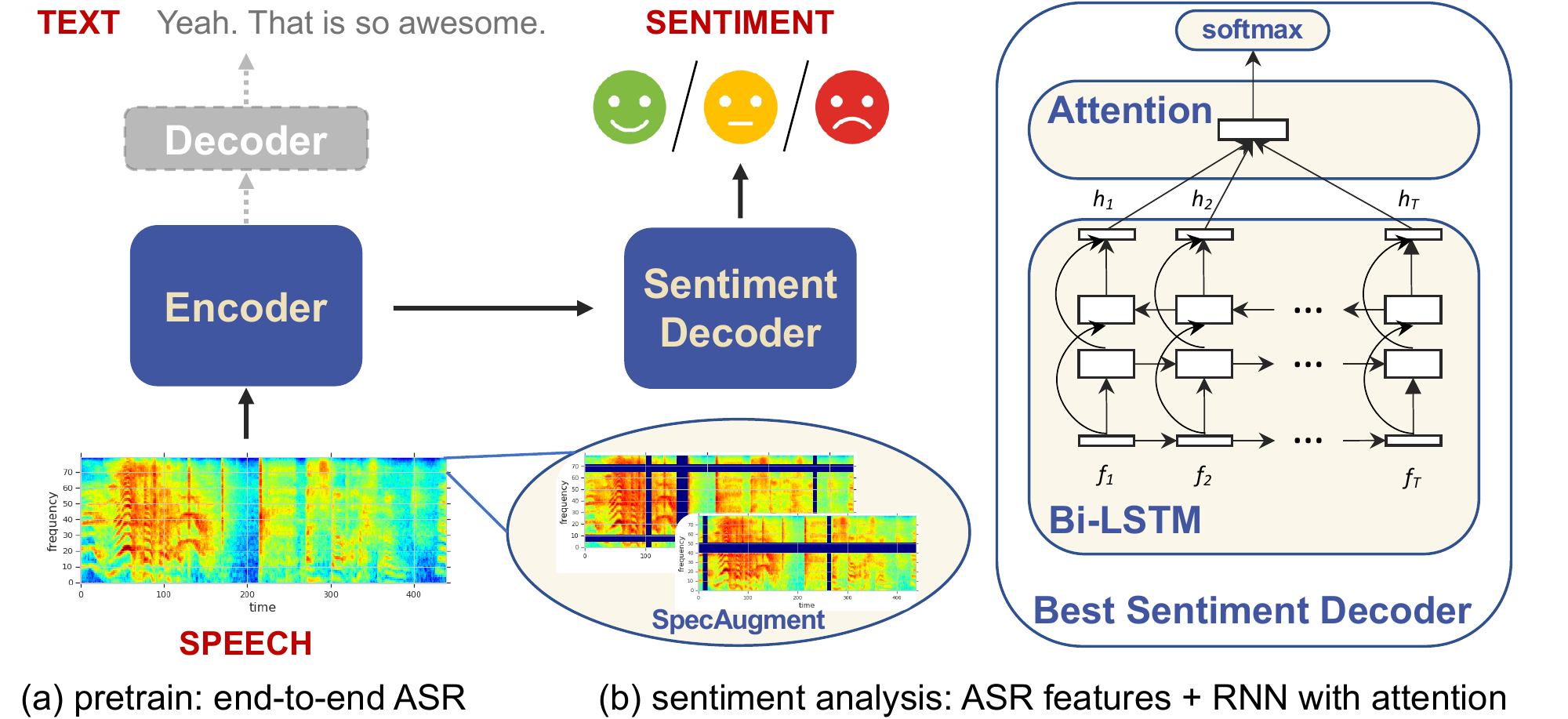}
\caption{We propose to use pre-trained features from e2e ASR model to solve sentiment analysis. The best performed sentiment decoder is RNN with self-attention. We apply SpecAugment to reduce overfitting in the training.  
\eat{E2e ASR features both acoustic and text information from speech, and can leverage
E2e ASR model automatically fuses linguistic and acoustic features in its hidden representations, which contains rich sentiment signals.}} \label{diagram}
\vspace{-0.1 in}
\end{figure}

\section{APPROACH}
\label{sec:approach}

In this section, we introduce our method to use ASR features for sentiment analysis. We first describe the end-to-end ASR model, and then introduce how we use ASR features for sentiment task. Lastly, we describe SpecAugment, a technique we use to reduce overfitting in training.
\eat{, and why it can be used to solve the sentiment classification task. }

Assume the input speech waveform has been transformed into a sequence of vectors, for example the log-mel features which we denote as $\vx_{1:T} = (x_1, x_2,\ldots, x_T )$. For sentiment task, the output is a label for the sequence, denoted as $y$.
\eat{\mbox{$y \in \{1,{.\hfil.\hfil.}, C\}$} where $C$ is the number of sentiment classes.}

\vspace{-0.1 in}
\subsection{End-to-end ASR Models}
Automatic Speech Recognition (ASR) is the problem of finding the most likely word sequence given a speech sequence. Recently end-to-end models have produced state-of-the-art performances in ASR tasks~\cite{prabhavalkar2017comparison}, \emph{without a separate pronunciation or language model}. 
In this work, we focus on one type of e2e model, RNN Transducer (\RNNT)\cite{graves2012sequence}, due to its simplicity.

Let $\vz_{1:L}$ be the output word sequence in ASR, where $z_l \in \mathcal{Z}$ graphemes. \eat{\cite{??}. word-piece tokens \cite{schuster2012japanese}.} \RNNT predicts the output by an encoder-decoder model. 
The encoder maps the input $\vx_{1:T}$ to a hidden representation $\vf_{1:T}$ \eat{= (f_1, \ldots, f_T )}, while the decoder maps the hidden representation $\vf_{1:T}$ to the text $\vz_{1:L}$.

\eat{is recurrent over the output sequence, conditioned on the hidden $\vf_{1:T}$.
\begin{align*}
&\vf_{1:T} = \text{Encoder}(\vx_{1:T}) & \text{representation learning}\\
& z_l  = \text{Decoder}(z_{l-1}, \vf_{1:T}
 ), \, \forall l. & \text{language model, alignment}
\end{align*}
}
The encoder is analogous to the \emph{acoustic model} in traditional ASR systems, where information in the speech is encoded into a higher-level vector representation. The decoder serves two purposes: first, it functions as the language model in traditional ASR. Secondly, it computes a distribution over all possible input-output alignments~\cite{graves2012sequence}. 
In practice, the language model module is called prediction network, and the alignment module is called joint network in \RNNT~\cite{prabhavalkar2017comparison}.
The encoder-decoder network can be jointly trained end-to-end. Details on the training and inference algorithm can be found in~\cite{graves2013speech}.

The encoder can be viewed as a feature extractor, which encodes useful information in the speech into $\vf_{1:T}$.
We hypothesize that $\vf_{1:T}$ contains rich sentiment signal, as it preserves both linguistic and acoustic characteristics.
Next we will describe how we use the ASR encoder features to classify sentiment.

\subsection{Sentiment Decoder on ASR Features}
\label{sec:sentidec}

Fig.~\ref{diagram} demonstrates the proposed framework. We take the encoder from the end-to-end ASR system, and freeze the encoder weights as a feature extractor. Note ASR feature $\vf_{1:T}$ is a sequence with variable length. 
Our sentiment classifier first transforms $\vf_{1:T}$ into a fixed length embedding, and the embedding is then fed into a softmax classifier to predict the sentiment label $y$. We call the process of mapping ASR features into fix length embedding as \emph{sentiment decoder}. The sentiment decoder is trained with cross-entropy loss. In what follows, we compare three sentiment decoders. 

\noindent \textbf{MLP with pooling.}
 A straightforward strategy is to build a multi-layer perceptron (MLP) on  $\vf_{1:T}$, and apply average (max) pooling over time to convert to a fixed-length representation. 

\noindent \textbf{RNN with pooling.}
However, MLP treats each time step in the sequence independently, which is not ideal, as sentiment depends on the context.
RNN is a prominent tool in modeling sequential data and is particularly good at storing and accessing context information over a sequence. \eat{modeling temporal dynamics}
We feed $\vf_{1:T}$ into a bi-directional LSTM (bi-LSTM), and concatenate the outputs of both forward and backward LSTMs into $\vh_{1:T}$.
\eat{:
\begin{align*}
\overrightarrow{h_t}  = \text{LSTM}_{\text{fwd}}(f_t, \overrightarrow{h_{t-1}}), \ \overleftarrow{h_t} = \text{LSTM}_{\text{bwd}}(f_t, \overleftarrow{h_{t+1}}).
\end{align*}

We concatenate the outputs of both LSTMs and get $h_t = [\overrightarrow{h_t}, \overleftarrow{h_t}]$ for each time step, which depends on the whole sequence.
Running LSTM over the sequence yields $\vh_{1:T}$. Finally}
We can either take the last hidden state $h_T$, or the average (max) pooling over $\vh_{1:T}$, as the sequence embedding. \eat{ to input to softmax.}

\noindent \textbf{RNN with multi-head self-attention.} 
Self-attention~\cite{lin2017structured} is good at handling long-term dependencies, which is critical for the sentiment task.
\eat{Learning long-term dependencies is yet a challenge for RNN, 
and we argue that global context is not strictly necessary when predicting sentiment. In fact, sentiment can be inferred from  local contexts.}
To this end, we propose to use a multi-head self-attention layer to replace the pooling or last hidden state, on top of RNNs. The multi-head attention layer can jointly attend to different subspaces of the input representation.
Furthermore, the attention alleviates the burden of learning long-term dependencies in LSTM via direct access to hidden states of the whole sequence~\cite{vaswani2017attention}. As a convenient side product, the soft attention weight provides an easy-to-visualize interpretation for the prediction.

Concretely, the multi-head self-attention computes multiple weighted sum of the LSTM hidden states, where the weights are given by soft attention vectors. Assume the attention layer has $\mathsf{n_a}$ heads with $\mathsf{d_a}$ dimensions for each head. $\mathsf{n_a}$ and $\mathsf{d_a}$ are hyper-parameters of our choice. The input to the attention head are the bi-LSTM hidden states $\vh_{1:T} \in \R^{T \times \mathsf{d_k}}$, which has length $T$ and dimension $\mathsf{d_k}$.
{\medmuskip=2mu
\thinmuskip=3mu
\thickmuskip=4mu
The attention vector $\va^i$ and the output $\vv^i$ of the $i$-th head is computed as
\begin{align*}
    \va^i = \text{softmax}\left(\frac{\vw^i_{\text{Q}} ( \vw^i_{\text{K}} \vh_{1:T} \T) }{\sqrt{\mathsf{d_a}}} \right), \quad \vv^i = \vw^i_{\text{V}} \vh_{1:T}\T \va^i,
\end{align*}
where the query token $\vw^i_{\text{Q}} \in \R^{\mathsf{1 \times d_a}}$, the key projection matrix $\vw^i_{\text{K}}\in \R^{\mathsf{d_a} \times \mathsf{d_k}}$ and the value projection matrix $\vw^i_{\text{V}}\in \R^{\mathsf{d_a} \times \mathsf{d_k}}$, are learnable parameters. $\va^i\in \R^T$ is the scaled dot-product attention~\cite{vaswani2017attention} probability. The weighted sum of hidden states $ \vv^i \in \R^{\mathsf{d_a}}$ is output. By concatenating the outputs from $\mathsf{n_a}$ heads, we obtain a fixed size embedding of length ${\mathsf{n_a} \times \mathsf{d_a}}$, which is then fed into softmax to classify sentiment.
}
\eat{
for $i=1, \ldots, \mathsf{n_a}$
Different from previous approaches, where either the final hidden state or the max (or average) pooling of hidden states of LSTM is used as the fixed-length embedding to feed into the classifier, the self-attention mechanism allows us to extract different aspects of the utterance into multiple vector representations through multi-heads. 
As a side product, the soft attention weight also provides us an interpretation for the prediction.
}

\subsection{Spectrogram Augmentation} \label{sec:specaug}
To reduce the overfitting in training sentiment classifiers, we employ Spectrogram Augmentation (SpecAugment)~\cite{park2019specaugment}, a data augmentation technique for speech. At training time, SpecAugment applies random transformations to the input, namely warping and masking on frequency channels or time steps, while keeps the label unchanged. Sentiment decoders trained with SpecAugment are invariant to small variations in the acoustic features, which improves the generalization. 
\eat{warps the acoustic
features, masks blocks of frequency channels, and masks blocks of time steps on filter bank coefficients. It is equivalent to creating more examples of the same label but with slightly different acoustic features.}

\eat{Our results show that SpecAugment is very easy to use and can effectively improve the performance of speech sentiment classification.}





\section{EXPERIMENT}
\label{sec:exp}

In this section, we first describe the  experimental setup in Section~\ref{subsec:setup}, and then show that our method using pre-trained ASR features can help sentiment analysis in Section~\ref{subsec:result}. We also examine the contribution of different components in Section~\ref{subsec:ablation}.
Lastly in Section~\ref{subsec:vis}, we visualize the attention weights to interpret the predictions, which can shed light on why ASR features can benefit the sentiment task. 
\subsection{Experiment Setup}
\label{subsec:setup}
\noindent \textbf{Data.\ } 
We use two datasets IEMOCAP and SWBD-sentiment in the experiments.
IEMOCAP~\cite{busso2008iemocap} is a well-benchmarked speech emotion recognition dataset. It contains approximately 12 hours audiovisual recording of both scripted and improvised interactions performed by actors\footnote{We use the speech data only in the experiments. We do not use the human annotated transcript.}. Following the protocol in~\cite{li2019dilated, kim2019dnn, cho2018deep,gu2018multimodal}, we experiment on a subset of the data, which contains 4 emotion classes \{happy+excited, neutral, sad, angry\}, with \{1708, 1084, 1636, 1103\} utterances respectively. We report 10-fold (leave-one-speaker-out) cross validation results. 

To further investigate speech sentiment task, we annotate a subset of switchboard telephone conversations~\cite{godfrey1997switchboard} with three sentiment labels, \ie negative, neutral and positive, and create the SWBD-sentiment dataset. SWBD-sentiment has over 140 hours of speech which contains approximately 49.5k utterances. We refer interested readers to~\cite{SWBDsenti2020} for more details about the dataset.
We split 10\% of SWBD-sentiment into a holdout set and a test set with 5\% each. We report the accuracy on the test set. 

Table~\ref{tab:data} provides a summary and comparison of two datasets.
Emotional expressions in IEMOCAP are elicited through acting in hypothetical scenarios, while sentiment in SWBD-sentiment is from natural conversations between friends. As a result, the class distribution in SWBD-sentiment is lightly imbalanced, with neutral, positive, and negative take up 52.6\%, 30.4\%, and 17.0\% respectively. 
Besides, utterances in SWBD-sentiment are generally longer. 
SWBD-sentiment is more challenging than IEMOCAP, but closer to real applications. 

\begin{table}[bthp]
    \centering
    \vspace{-.2in}
    \caption{Speech sentiment datasets} \label{tab:data}
    \begin{adjustbox}{width=0.49 \textwidth}
    \begin{tabular}{c|c|c|c}
    \hline
       dataset  & \# utterances / hours & \# classes & elicitation \\ \hline
       IEMOCAP  & 10k / 12h  & 4 & play-acted\\ \hline
       SWBD-sentiment & 49.5k / 140h & 3 & conversation \\ \hline
    \end{tabular}
    \end{adjustbox}
    \label{tab:my_label}
\end{table}

\noindent\textbf{Metrics.\ } 
We report both weighted accuracy \WA and unweighted accuracy \UA, commonly used for speech sentiment analysis~\cite{kim2019dnn,gu2018multimodal}. \WA is the conventional classification accuracy, \ie the percentage of samples correctly labeled by the model. 
To sidestep the effect of class imbalance on \WA, we also report \UA, the average accuracy of different classes.

\noindent\textbf{Baselines.\ }
We compare the proposed approach to both single-modality and multi-modality models. A single-modality model uses audio inputs only, while a multi-modality model uses both audio and ASR transcription as inputs. On IEMOCAP, we use the state-of-the-art results\cite{li2019dilated,kim2019dnn} from both approaches reported in the literature as baselines. 
On SWBD-sentiment, we implemented our own baselines. We thoroughly tuned the architectures from MLPs, CNNs, LSTMs, and a combination of them. We tuned the number of layers from $\{2, 3, 4, 5\}$, and tuned the filter sizes and strides for CNNs, and number of hidden units for LSTMs and MLPs. We report the best results from tuning.

\noindent\textbf{Model architecture.\ }
\label{subsec:implent}
All experiments use 80-dimensional features, computed with
a 25ms window and shifted every 10ms. 
We use a pre-trained \RNNT model trained from  YouTube videos 
described in \cite{soltau2017neural}. The encoder stacks a macro layer $3$ times, where the macro layer consists of $512$ $1$-D convolutions with filter width $5$ and stride $1$, a $1$-D max pooling layer with width $2$ and stride $2$, and $3$ bidirectional LSTM layers with $512$ hidden units on each direction and a $1536$-dimensional projection per layer.  The prediction network has a unidirectional LSTM with $1024$ hidden units. The joint network has $512$ hidden units and the final output use graphemes.

For the sentiment classifier, we use 1 bidirectional LSTM layer with 64 hidden units on each direction, and 1 multi-head self-attention, which has 8 heads with 32 units per head. 
For other classifiers in Section~\ref{subsec:ablation}, we use respective layers matching the number of hidden nodes. 
The SpecAugment parameters are the same as the LibriSpeech basic (LB) policy of Table 1 in~\cite{park2019specaugment}.
All models are trained with Adam optimizer in the open-source Lingvo toolkit~\cite{shen2019lingvo} with learning rate $10^{-4}$, and gradient clipping norm $4.0$.

\begin{table*}[!bhtp]
    \centering
    \caption{Speech sentiment analysis performances of different methods. \eat{I feel it is a bit confusing to have audio + ASR text. Maybe we can remove this row and put it into another table (say audio + predicted transcript vs end2end ASR?)}}\label{tab:mainres}
    \begin{tabular}{|c| c c c| c c c|}
    \hline
    & \multicolumn{3}{c|}{IEMOCAP dataset} & \multicolumn{3}{c|}{SWBD-sentiment dataset} \\
    \hline 
         Input features &  Architecture   & \WA (\%) & \UA (\%)  &  Architecture   & \WA (\%) & \UA (\%)\\
        \hline
      acoustic &  DRN + Transformer~\cite{li2019dilated}  &  -  & 67.4 & CNN & 54.23 & 39.63\\
      acoustic + text &  DNN ~\cite{kim2019dnn} &  66.6  & 68.7 & CNN and LSTM & 65.65 & 54.59\\
        \hline 
        e2e ASR & RNN w/ attention & 71.7 & 72.6 & RNN w/ attention & 70.10 & 62.39 \\
    \hline \hline
     - & human &  91.0 & 91.2 & human & 85.76 & 84.61\\ \hline
    \end{tabular}
\end{table*}
\subsection{Sentiment Classification}
\label{subsec:result}
Table~\ref{tab:mainres} summarizes the main results of our approach and state-of-the-art methods on IEMOCAP and SWBD-sentiment. We provide a reference of the human performance, which is essentially the average agreement percentage among annotators.
Since sentiment evaluation is subjective, human performance is an upper bound on the accuracies. Comparing human performance on SWBD-sentiment with IEMOCAP, we confirm that SWBD-sentiment is more challenging.

On IEMOCAP, ASR features with RNN and self-attention decoder achieves 71.7\% accuracy (\WA), which improves the state-of-the-art by 5.1\%. 
On SWBD-sentiment, a naive baseline to predict the neutral class for all samples can achieve 52.6\% \WA and 33.3\% \UA. Training deep models directly on acoustic features suffers from severe ovefitting, and only improves over this baseline by 1.6\% on \WA. Fusing acoustic features with text can significantly improve the performance to 65.65\%. Sentiment decoder with e2e ASR features achieves the best 70.10\% accuracy. Experiments on both datasets demonstrate that our method can improve speech sentiment prediction. 

\eat{computed as
by treating each annotation as a prediction and comparing it with the assigned label, the label the most agree on. \TODO{I don't understand this description. Is it just inter annotator disagreement? The description sounds "ground truth" is not an annotation.}}

\vspace{-.1in}
\subsection{Ablation Study}
\label{subsec:ablation}
In this section, we study the contribution of different components in the proposed method. We compare the performances of different sentiment decoders, and analyze the effect of SpecAugment.

In table~\ref{tab:ablation}, the first row is our best performing model, which trains a bi-LSTM with attention decoder on ASR features using SpecAugment. We refer to it as the base model.
The second and third rows are based on different sentiment decoders, MLP with pooling and RNN with pooling respectively.  RNN with attention is better than RNN with pooling while RNN with pooling is better than MLP with pooling. Somewhat surprisingly, a simple MLP with ASR features achieves 68.68\% accuracy which already improves over the state-of-the-art 66.6\% in the literature. This demonstrates the effectiveness of  pre-trained ASR models.
The last row reports the accuracy of the base model when trained without SpecAugment, which is roughly 1\% worse. In all our experiments, we find SpecAugment consistently helpful on different sentiment decoders and datasets. 
\begin{table}[bhtp]
    \centering
    \vspace{-0.1 in}
    \caption{Ablation study on IEMOCAP dataset}\label{tab:ablation}
    \begin{tabular}{l|l|c|c}
    \hline
      \multicolumn{2}{c|}{  description}   & \WA (\%) & \UA (\%) \\
        \hline 
        \multicolumn{2}{l|}{ RNN w/ attention + SpecAugment } & 71.72 & 72.56 \\ \hline \hline
        \multirow{2}{*}{decoders}& MLP w/ pooling & 68.68 & 68.98 \\
        & RNN w/ pooling & 70.71 & 71.55\\
        \eat{&  self-attn. only &  & \\}
        \hline
         \multicolumn{2}{l|} {w/o SpecAugment}  & 70.77 & 71.77\\ \hline 
    \end{tabular}
    \vspace{-0.05 in}
\end{table}

\vspace{-.2in}
\subsection{Attention Visualization}
\label{subsec:vis}
\begin {figure}[thbp]
\centering
    \includegraphics[width=0.49\textwidth]{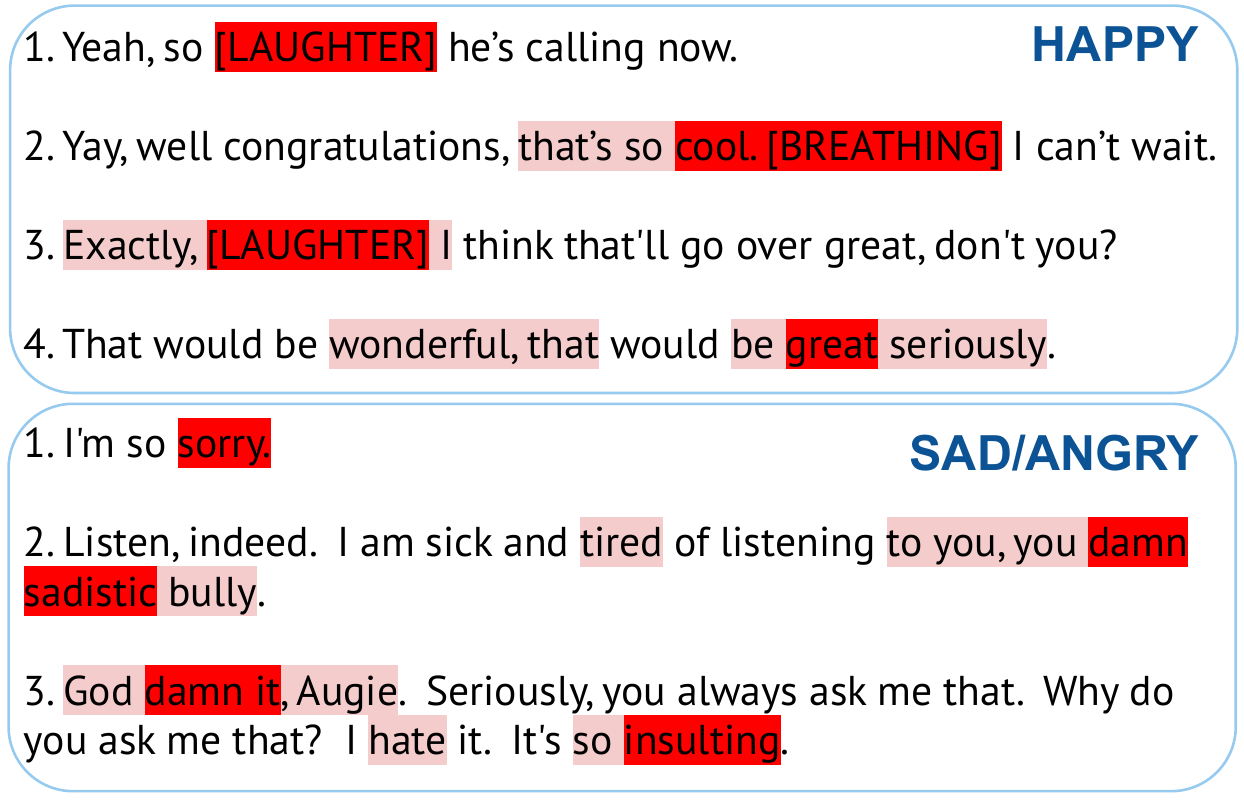}
\caption{Attention visualization on IEMOCAP utterances from different sentiment classes. There are two notable patterns that \eat{are likely to }have larger weights: specific vocalizations, like [LAUGHTER] and [BREATHING], and indicating words, like ``great'' and ``damn''. This supports our hypothesis that ASR features contain both acoustic and text information.} \label{fig:viz}
\vspace{-0.1 in}
\end{figure}

We interpret the predictions by examining the attention $\va^i$ (see Section~\ref{sec:sentidec}) on utterances. 
$\va^i$ is of the same length as ASR features, and its element indicates the contribution of each frame to the prediction. 
However, it is hard to illustrate attention over audio on paper. Instead, we visualize attention over ASR transcripts, by running alignment and taking average attention of frames aligned to one word as its weight.
We add tokens, like [LAUGHTER] or [BREATHING], to the ASR text to annotate non-verbal vocalizations for visualization purpose.
We quantize the attention weights into three bins, and draw a heat map in Fig.~\ref{fig:viz}. The visualization result demonstrates how our model integrates both acoustic and language features to solve sentiment analysis.

\eat{\llcao{We shall clarify that [laugh] or [breathe] does appears in the transcribed text. Otherwise the reviewers may  perceive wrongly. }}

\eat{
We also find that the attention weights for longer sequences are less likely to be sparse and peaky, compared to shorter ones. And several heads can provide similar attentions. A possible solution is to add sparsity regularizer and diversity penalty among heads to the loss, which remains an interesting future direction. 
}

\eat{Despite little tuning of hyperparameters, this simple model achieves excellent
results on multiple benchmarks, suggesting that
the pre-trained ASR features are `universal' feature extractors that can be utilized for sentiment classification tasks.}

\vspace{-.1in}
\section{CONCLUSION AND FUTURE WORK}
\label{sec:conclusion}
In this paper, we demonstrate pre-trained features from the end-to-end ASR model are effective on sentiment analysis. It improves the-state-of-the-art accuracy on IEMOCAP dataset from 66.6\% to 71.7\%. Moreover, we create a large-scale speech sentiment dataset SWBD-sentiment to facilitate future research in this field. 
Our future work includes experimenting with unsupervised learnt speech features, as well as applying end-to-end ASR features to other down-stream tasks like diarization, speaker identification, and etc. 


\eat{
End-to-end ASR feature as general purpose speech representation, and ASR as universal pre-training.}

\vspace{.1in}
\noindent \textbf{Acknowledgment} \hspace{.2in}
We are grateful to Rohit Prabhavalkar, Ruoming Pang, Wei Han, Bo Li, Gary Wang, and Shuyuan Zhang for their fruitful discussions and suggestions.
\small{
\bibliographystyle{IEEEbib}
\bibliography{refs}
}
\end{document}